\documentclass[acmtog]{acmart}
\acmSubmissionID{1269}

\usepackage{graphicx}
\usepackage{amsmath}
\usepackage{booktabs}
\usepackage{mathtools}
\usepackage{multirow}
\usepackage{caption}
\usepackage{adjustbox}
\usepackage{float}
\usepackage{enumitem}

\usepackage[capitalize]{cleveref}
\crefname{section}{Sec.}{Secs.}
\Crefname{section}{Section}{Sections}
\Crefname{table}{Table}{Tables}
\crefname{table}{Tab.}{Tabs.}

\newcommand{\adjective} {authentic}
\newcommand{\Adjective} {Authentic}

\AtBeginDocument{%
  }

\setcopyright{rightsretained}
\acmJournal{TOG}
\acmYear{2024} \acmVolume{43} \acmNumber{4} \acmArticle{135} \acmMonth{7}\acmDOI{10.1145/3658237}

\begin{document}

\title{RealFill: Reference-Driven Generation for Authentic Image Completion}

\author{Luming Tang}
\email{lt453@cornell.edu}
\affiliation{
  \institution{Cornell University}
  \country{US}
}

\author{Nataniel Ruiz}
\email{natanielruiz@google.com}
\affiliation{
  \institution{Google Research}
  \country{US}
}

\author{Qinghao Chu}
\email{qchu@google.com}
\affiliation{
  \institution{Google Research}
  \country{US}
}

\author{Yuanzhen Li}
\email{yzli@google.com}
\affiliation{
  \institution{Google Research}
  \country{US}
}

\author{Aleksander Hołyński}
\email{holynski@google.com}
\affiliation{
  \institution{Google Research}
  \country{US}
}

\author{David E. Jacobs}
\email{dejacobs@google.com}
\affiliation{
  \institution{Google Research}
  \country{US}
}

\author{Bharath Hariharan}
\email{bharathh@cs.cornell.edu}
\affiliation{
  \institution{Cornell University}
  \country{US}
}

\author{Yael Pritch}
\email{yaelp@google.com}
\affiliation{
  \institution{Google Research}
  \country{Israel}
}

\author{Neal Wadhwa}
\email{nealw@google.com}
\affiliation{
  \institution{Google Research}
  \country{US}
}

\author{Kfir Aberman}
\email{kfiraberman@gmail.com}
\affiliation{
  \institution{Snap Research}
  \country{US}
}

\author{Michael Rubinstein}
\email{mrub@google.com}
\affiliation{
  \institution{Google Research}
  \country{US}
}

\renewcommand{\shortauthors}{Tang et al.}

\begin{abstract}
Recent advances in generative imagery have brought forth outpainting and inpainting models that can produce high-quality, plausible image content in unknown regions. However, the content these models hallucinate is necessarily inauthentic, since they are unaware of the true scene. In this work, we propose \textbf{RealFill}, a novel generative approach for image completion that fills in missing regions of an image with the content that \underline{should have been there}. RealFill is a generative inpainting model that is personalized using only a few reference images of a scene. These reference images do not have to be aligned with the target image, and can be taken with drastically varying viewpoints, lighting conditions, camera apertures, or image styles. Once personalized, RealFill is able to complete a target image with visually compelling contents that are faithful to the original scene. We evaluate RealFill on a new image completion benchmark that covers a set of diverse and challenging scenarios, and find that it outperforms existing approaches by a large margin.
Project page: \url{https://realfill.github.io}.
\end{abstract}



\begin{teaserfigure}
  \includegraphics[width=\textwidth]{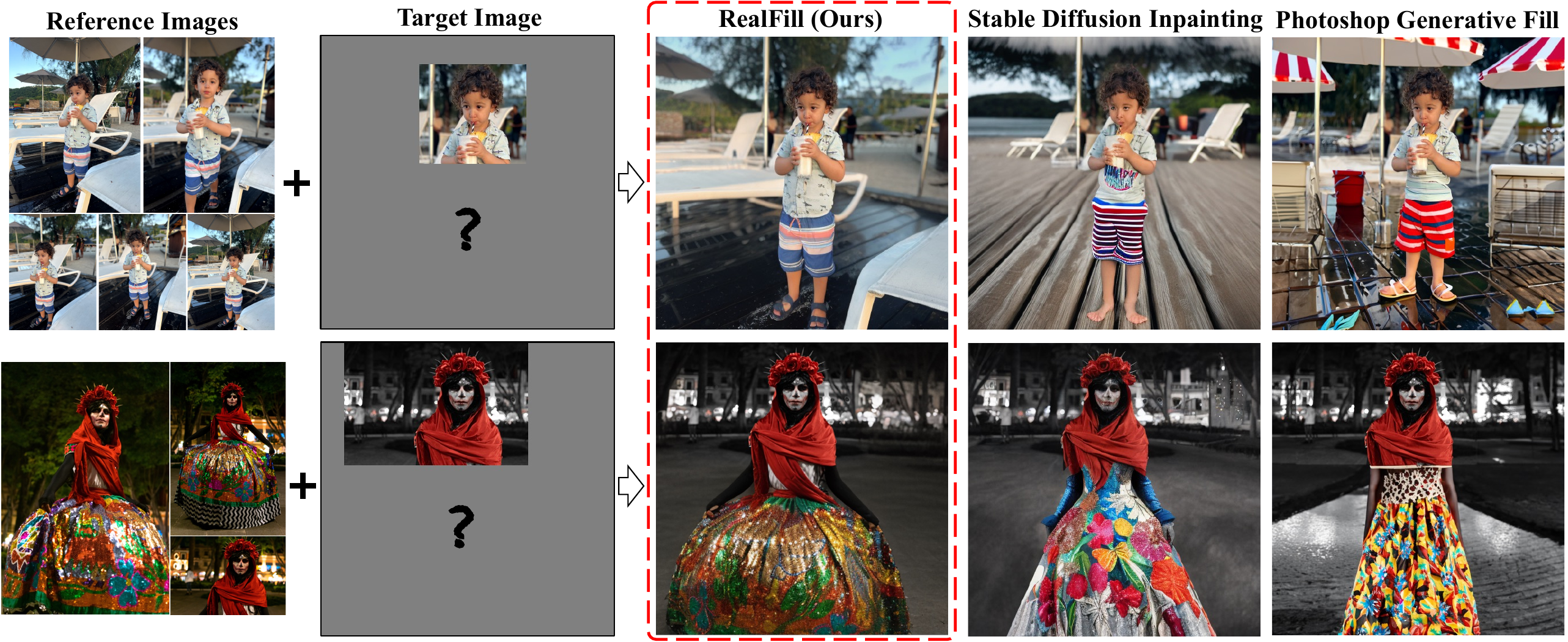}
  \vspace{-15pt}
  \caption{Given a few reference images that roughly capture the same scene, and a target image with a missing region, \textit{\textbf{RealFill}} is able to complete the target image with image content that is faithful to the true scene. In contrast, standard prompt-based inpainting methods hallucinate plausible but inauthentic content due to their lack of knowledge of the original scene.}
  \label{fig:teaser}
\end{teaserfigure}

\maketitle

\section{Introduction}
Photographs capture ephemeral and invaluable experiences in our lives, but can sometimes fail to do these memories justice. In many cases, no single shot may have captured the perfect angle, framing, timing, or composition, and unfortunately, just as the experiences themselves cannot be revisited, these elements of the captured images are also unalterable. 
We show one such example in Fig.~\ref{fig:method}: imagine having taken a nearly perfect photo of your daughter dancing on stage, but her unique and intricate crown is partially cut out of the frame. Of course, there are many other pictures from the performance that showcase her crown, but they all fail to capture that precise special moment: her pose mid-dance, her facial expression, and the perfect lighting. Given this collection of imperfect photos, you can certainly imagine the missing parts of this perfect shot, but actually creating a complete, shareable version of this image is much harder.

In this paper, we focus on this problem, which we call \textit{\Adjective{} Image Completion}. Given a few reference images (up to five) and one target image that captures roughly the same scene (but in a different arrangement or appearance), we aim to fill missing regions of the target image with high-quality image content that is faithful to the originally captured scene. Note that for the sake of practical benefit, we focus particularly on the more challenging, unconstrained setting in which the target and reference images may have very different viewpoints, environmental conditions, camera apertures, image styles, or even moving objects.

Approaches to solving variants of this problem have been proposed using classical geometry-based pipelines~\cite{shan2014photo, zhou2021transfill, zhao2023geofill} that rely on correspondence matching, depth estimation, and 3D transformations, followed by patch fusion and image harmonization. These methods tend to encounter catastrophic failure when the scene's structure cannot be accurately estimated, e.g., when the scene geometry is too complex or contains dynamic objects. 
On the other hand, recent generative models~\cite{yu2018generative, chang2022maskgit, chang2023muse}, and in particular diffusion models~\cite{ho2020denoising, song2021scorebased, rombach2022high}, have demonstrated strong performance on the tasks of image inpainting and outpainting~\cite{ramesh2022hierarchical, sd2inpaint, wang2023imagen}. These methods, however, struggle to recover genuine scene structure and fine details, since they are only guided by text prompts, and can't use reference image content. 

To this end, we present a simple yet effective reference-driven image completion framework called \textit{RealFill}. For a given scene, we first create a personalized generative model by finetuning a pretrained inpainting diffusion model~\cite{sd2inpaint} on the reference and target images. This finetuning process is designed such that the adapted model not only maintains a good image prior, but also learns the contents, lighting, and style of the scene in the input images. 
We then use this finetuned model to fill the missing regions in the target image through a standard diffusion sampling process.
Given the stochastic nature of generative inference, we 
propose \textit{Correspondence-Based Seed Selection}, to automatically select a small set of high-quality generations by exploiting a special property of our completion task: that there exists true correspondences between generated contents and reference images. Specifically, we filter out samples that have too few keypoint matches with references, which greatly reduces human labor to select high-quality model outputs.

As shown in \cref{fig:teaser,fig:method,fig:gallery,fig:gallery_inpaint}, RealFill is able to effectively inpaint and outpaint a target image with its \textit{genuine} scene content. Most importantly, our method is able to handle large differences between reference and target images, e.g., viewpoint, lighting, aperture, style or dynamic deformations --- differences which are very difficult for previous geometry-based approaches. 
Existing benchmarks for image completion~\cite{zhou2021transfill} mainly focus on small inpainting tasks and minimal changes between reference and target images.
In order to quantitatively evaluate the aforementioned challenging use-case, we collect a dataset containing 10 inpainting and 23 outpainting examples along with corresponding ground-truth, and show that RealFill outperforms baselines by a large margin across multiple image similarity metrics.

In summary, our contributions are as follows:
\begin{itemize}
    \item We define a new problem named \textit{\Adjective{} Image Completion}, i.e., given a set of reference images and a target image with missing regions, we seek to complete those missing regions with content that is faithful to the scene as captured in the references. 
    In essence, the goal is to complete the target image with what ``should have been there'' rather than what ``could have been there'', like in typical generative inpainting.
    \item We introduce \textit{RealFill}, a method that aims to solve this problem by finetuning an inpainting diffusion model on reference and target images. This model is sampled with \textit{Correspondence-Based Seed Selection} to filter outputs with low fidelity to the reference images. \textit{RealFill} is the first method that expands the expressive power of generative inpainting models by conditioning the process beyond text, enabling extra conditioning on reference images.
    \item We propose \textit{RealBench}, a dataset for quantitative evaluation of \adjective{} image completion, composed of 33 scenes spanning both inpainting and outpainting tasks. 
\end{itemize}
\section{Related Work}

\textbf{Adapting Pretrained Diffusion Models}.
Diffusion models~\cite{ho2020denoising, song2021scorebased, dhariwal2021diffusion} have shown strong performance in text-to-image (T2I) generation~\cite{ramesh2022hierarchical, saharia2022photorealistic, rombach2022high}. Recent works make use of this pretrained image prior by finetuning them for various tasks.
Personalization methods propose to finetune the T2I model~\cite{ruiz2023dreambooth, ruiz2023hyperdreambooth, chen2023subject, avrahami2023break} or text embedding~\cite{gal2022image,voynov2023p+}, on a few images to achieve arbitrary text-driven generation of a given object or style. 
Other techniques instead finetune a T2I model to add new conditioning signals, either for image editing~\cite{kawar2023imagic, brooks2023instructpix2pix, wang2023imagen} or more controllable generation~\cite{zhang2023adding, mou2023t2i, sohn2023styledrop}.
The same approach can be also applied to specialized tasks~\cite{liu2023zero1to3, raj2023dreambooth3d, wu2022tuneavideo, zhao2023unleashing} such as converting a T2I model into a 3D or video generation model.
Our method shows that a pretrained T2I inpainting diffusion model can be adapted to perform reference-driven image completion.

\noindent\textbf{Image Completion}. As an enduring challenge in computer vision, image completion aims to fill missing parts of an image with plausible content, i.e., inpainting and outpainting. Traditional approaches~\cite{bertalmio2000image, criminisi2003object, hays2007scene, barnes2009patchmatch} rely on handcrafted heuristics while more recent deep learning-based methods~\cite{iizuka2017globally, liu2018image, suvorov2022resolution, kim2022zoom} instead directly train end-to-end neural networks that take original image and mask as inputs and generate the completed image. Given the challenging nature of this problem~\cite{zheng2019pluralistic}, many works~\cite{yeh2017semantic, chang2022maskgit, lugmayr2022repaint, chang2023muse} propose to leverage the image prior from a pretrained generative model for this task. Built upon powerful T2I diffusion models, recent solutions~\cite{sd2inpaint, ramesh2022hierarchical, adobephotoshop} demonstrate strong text-driven image completion capabilities. However, due to their sole dependence on a text prompt (which has limited descriptive power), generated image content can often be hard to control, resulting in tedious prompt engineering, especially when a particular or otherwise true scene content is desired. This is one of the main issues we aim to tackle in our work.

\noindent\textbf{Reference-Based Image Inpainting}. 
Existing works for reference-based inpainting~\cite{zhou2021transfill, zhao2023geofill} or outpainting~\cite{shan2014photo} usually make use of carefully tuned pipelines containing many individual components like depth and pose estimation, image warping, and harmonization. Each of these modules usually tackles a moderately challenging problem itself and the resulting prediction error can, and often does, propagate and accumulate through the pipeline. This can lead to catastrophic failure especially in challenging cases with complex scene geometry, changes in appearance, or scene deformation. 
Paint-by-Example~\cite{yang2023paint} proposes a novel latent diffusion model~\cite{rombach2022high} whose generation is conditioned on both a reference and target image. However, the conditioning is based on CLIP embedding~\cite{radford2021learning} of a single reference image, therefore is only able to capture high-level semantics of the reference object. 
In contrast, our method is the first to demonstrate multiple reference image-driven inpainting and outpainting that is both visually compelling and faithful to the original scene, even in cases where there are large appearance changes between reference and target images.
\section{Method}

\begin{figure*}
\centering
\makebox[\textwidth][c]{\includegraphics[width=1\textwidth]{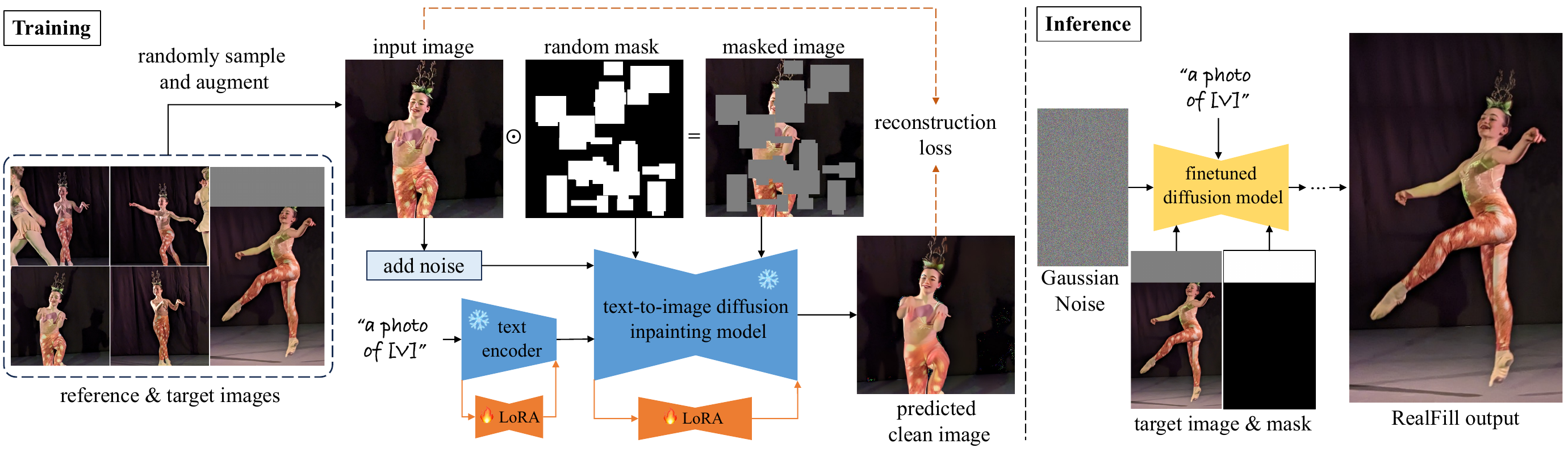}}
\vspace{-0.55cm}
\caption{
\textbf{Training and inference pipelines of RealFill}. RealFill's inputs are a target image to be filled and a few reference images of the same scene. We first finetune LoRA weights of a pretrained inpainting diffusion model on the reference and target images (with random patches masked out). Then, we use the adapted model to fill the desired region of the target image, resulting in a faithful, high-quality output. For example, the girl's crown is recovered in the target image, despite the girl being in very different poses in the reference images.
}
\label{fig:method}
\end{figure*}

\subsection{Reference-Based Image Completion}
\label{sec:approach}

Given a set of casually captured reference images (up to five), our goal is to complete (either outpaint or inpaint) a target image of roughly the same scene. The output image is expected to be not only plausible and photorealistic, but also faithful to the reference images, i.e., recovering content and scene detail that were present in the actual scene. In essence, we want to achieve \textit{\adjective{} image completion}, where we generate what ``should have been there'' instead of what ``could have been there''. We purposefully pose this as a broad and challenging problem with few constraints on the inputs. For example, the images could be taken from very different viewpoints with unknown camera poses. They could also have different lighting conditions or styles, and the scene could potentially be non-static and have significantly varying layout across images.

In this section, we first provide background knowledge on diffusion models and subject-driven generation (Sec.~\ref{sec:prelim}). Then, we formally define the problem of \adjective{} image completion (Sec.~\ref{sec:problem}). Finally, we present RealFill, our method to perform reference-based image completion with a pretrained diffusion image prior (Sec.~\ref{sec:realfill}).

\subsection{Preliminaries}
\label{sec:prelim}
\textbf{Diffusion models} are generative models that aim to transform a Normal distribution into an arbitrary target data distribution. During training, different magnitudes of Gaussian noise are added to a data point $x_0$ to obtain noisy $x_t$:
\begin{equation}
    x_{t} = \sqrt{\alpha_t} x_0 + (\sqrt{1-\alpha_t})\epsilon
\label{eq:forward}
\end{equation}
where the noise $\epsilon\sim\mathcal{N}(0, \mathbf{I})$, and $\{\alpha_t\}_{t=1}^{T}$ define a fixed noise schedule with larger $t$ corresponding to more noise. Then, a neural network $\epsilon_\theta$ is trained to predict the noise using the following loss function:
\begin{equation}
    \mathcal{L} = \mathbb{E}_{x, t, \epsilon}\left\|\mathbf{\epsilon}_\theta(x_t, t, c) - \mathbf{\epsilon}\right\|^2_2
\label{eq:loss}
\end{equation}
where $\epsilon_\theta$ is conditioned on some signal $c$, e.g., a language prompt for a T2I model, a masked image for an inpainting model. During inference, starting from $x_{T}\sim\mathcal{N}(0, \mathbf{I})$, $\epsilon_\theta$ is used to iteratively remove noise from $x_t$ to get a less noisy $x_{t-1}$, eventually leading to a sample 
$x_0$ from the target data 
distribution.

\noindent\textbf{DreamBooth}~\cite{ruiz2023dreambooth} enables T2I diffusion models to generate images of a specific subject with semantic modifications. The core idea is to finetune the model $\epsilon_\theta$ on a few subject images using the loss in Eq.~\ref{eq:loss}. Instead of finetuning all the network weights, it is possible to combine DreamBooth with Low Rank Adaptations (LoRA)~\cite{lora_stable, hu2022lora}, for a more memory-efficient alternative, by injecting learnable residual modules $\Delta W$ to each network weight matrix $W$. $\Delta W$ is a composition of low-rank matrices, i.e., $W+\Delta W=W+AB$ where $W\in\mathbb{R}^{n\times n}$, $A\in\mathbb{R}^{n\times r}$, $B\in\mathbb{R}^{r\times n}$, $r\ll n$, and only the added $\Delta W$ is being updated during training while model's original parameters $W$ stay frozen.

\subsection{Problem Setup}
\label{sec:problem}
Formally, the model is given $n$ ($n\le 5$) reference images $\mathcal{X}_{ref}\coloneqq\{I_{ref}^k\}_{k=1}^{n}$, 
a target image $I_{tgt}\in\mathbb{R}^{H\times W\times 3}$ and its associated binary mask $M_{tgt}\in\{0, 1\}^{H\times W}$, in which 1 denotes the region to fill and 0 denotes the existing area in $I_{tgt}$. The model is expected to generate a harmonized image $I_{out}\in\mathbb{R}^{H\times W\times 3}$ whose pixels should stay as similar as possible to $I_{tgt}$ where $M_{tgt}$ equals 0, while staying faithful to the corresponding contents in $\mathcal{X}_{ref}$ where $M_{tgt}$ equals 1. We assume there is enough overlap between $\mathcal{X}_{ref}$ and $I_{tgt}$ such that a human could imagine a plausible $I_{out}$.

\subsection{RealFill}
\label{sec:realfill}

This task is challenging for both geometry-based~\cite{zhou2021transfill, zhao2023geofill} and reconstruction-based approaches~\cite{mildenhall2020nerf} because there are barely any geometric constraints between $\mathcal{X}_{ref}$ and $I_{tgt}$, there are only a few images available as inputs, and the reference images may have different styles, lighting conditions, and subject poses from the target. One alternative is to use a controllable inpainting or outpainting model, however, these models are either prompt-based~\cite{rombach2022high, adobephotoshop} or single-image object-driven~\cite{yang2023paint}, which makes them hard to use for recovering complex scene-level structure and details.

Therefore, we propose to first inject knowledge of the scene into a pretrained generative model by finetuning it on a set of reference images, then use the adapted model to generate $I_{out}$ conditioned on $I_{tgt}$ and $M_{tgt}$, such that it is aware of the scene's contents.

\noindent\textbf{Training}.
Starting from a state-of-the-art T2I diffusion inpainting model~\cite{rombach2022high}, we inject LoRA weights and finetune it on both $\mathcal{X}_{ref}$ and $I_{tgt}$ with randomly generated binary masks $m\in\{0, 1\}^{H\times W}$. The loss function is
\begin{equation}
    \mathcal{L} = \mathbb{E}_{x, t, \epsilon, m}\left\|\mathbf{\epsilon}_\theta(x_t, t, p, m, (1-m)\odot x) - \mathbf{\epsilon}\right\|^2_2
\label{eq:inpaint-loss}
\end{equation}
where $x\in\mathcal{X}_{ref}\cup\{I_{tgt}\}$, $p$ is a fixed language prompt, $\odot$ denotes the element-wise product and therefore $(1-m)\odot x$ is the masked clean image. For $I_{tgt}$, the loss is only calculated on the existing region, i.e., where $M_{tgt}$'s entry equals 0. 
Specifically, we use the open-sourced Stable Diffusion v2 inpainting model~\cite{sd2inpaint} and inject LoRA layers into its text encoder and U-Net for finetuning. Following~\cite{ruiz2023dreambooth}, we fix $p$ to be a sentence containing a rare token, i.e., ``\texttt{a photo of [V]}". For each training example, similar to \cite{suvorov2022resolution}, we generate multiple random rectangles and take either their union or the complement of the union to get the final random mask $m$. 
Fig.~\ref{fig:method} illustrates the whole pipeline.

\noindent\textbf{Inference}.
After training, we use the DDPM~\cite{ho2020denoising} sampler to generate an image $I_{gen}$, conditioning the model on $p$, $I_{tgt}$ and $M_{tgt}$. 
However, similar to the observation in~\cite{zhu2023designing}, we notice that the existing region in $I_{tgt}$ is distorted in $I_{gen}$. To resolve this, we first feather the mask $M_{tgt}$, then use it to alpha composite $I_{gen}$ and $I_{tgt}$, leading to the final $I_{out}$ with full recovery on the existing area and a smooth transition at the boundary of the generated region.

\noindent\textbf{Correspondence-Based Seed Selection}.
The diffusion inference process is stochastic, i.e., the same inputs may produce any number of generated images depending on the random seeds of the sampling process.  
The generation quality can vary due to this stochasticity, thus requiring humans to select high-quality samples. While there is work to identify good samples from a collection of generated outputs~\cite{samuel2023all}, this remains an open problem. 
In our case, the reference images actually provide a grounding signal for the true scene content, thus can be used to help identify high-quality outputs.

Specifically, we use the number of image feature correspondences between $I_{out}$ and $\mathcal{X}_{ref}$ as a metric to roughly quantify whether the result is faithful to the reference images.
During inference, we first generate a batch of outputs, i.e., $\{I_{out}\}$, then extract a set of correspondences (e.g., using LoFTR~\cite{sun2021loftr}) between $\mathcal{X}_{ref}$ and the filled region of each $I_{out}$, and finally rank the generated results $\{I_{out}\}$ by the number of correspondences. 
This allows us to automatically filter generations to  a small set of high-quality results.

\section{Experiments}
\label{sec:exp}

\begin{figure*}
\centering
\makebox[\textwidth][c]{\includegraphics[width=\textwidth]{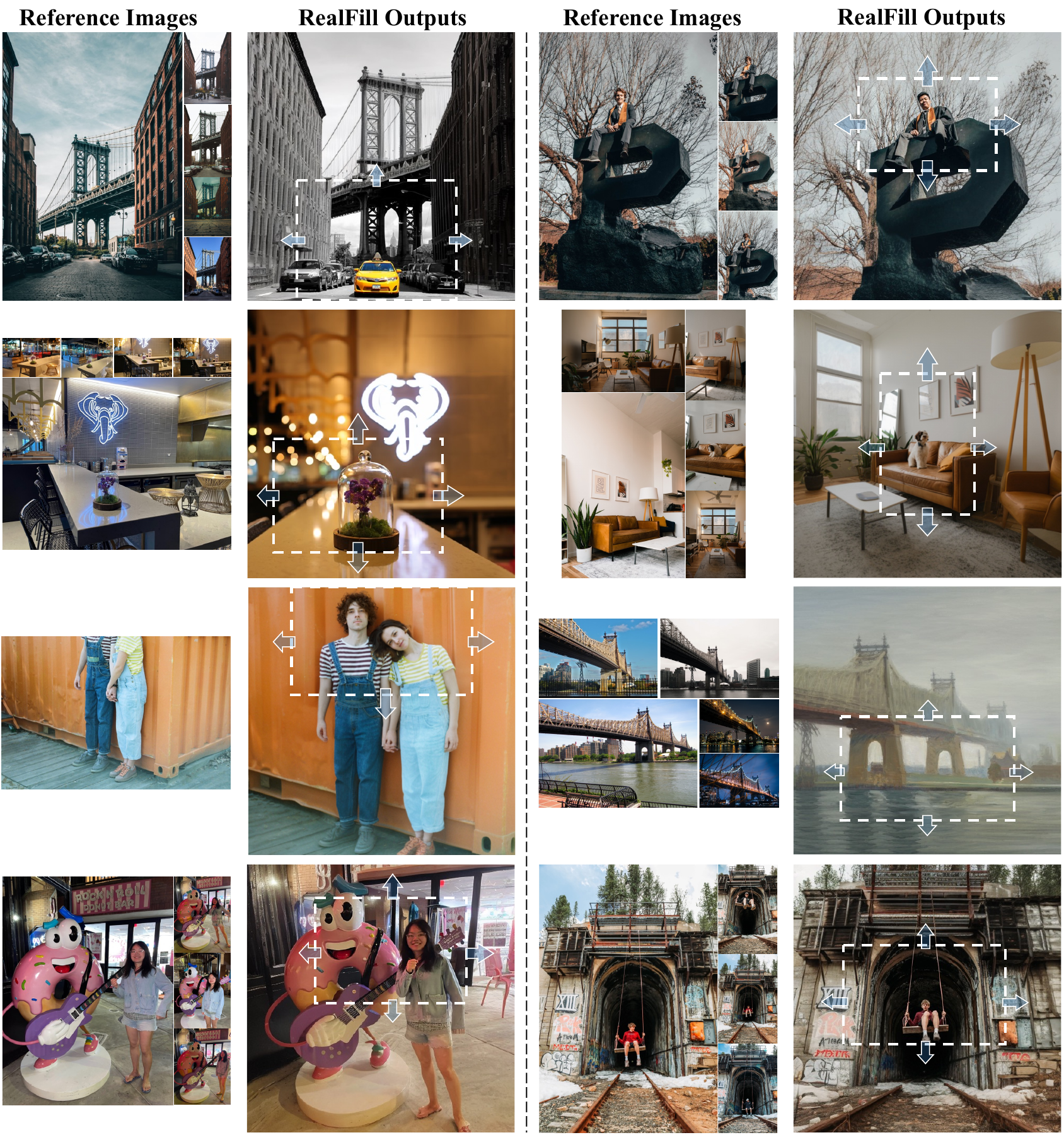}}
\caption{
\textbf{Reference-based outpainting with RealFill}. Given the reference images on the left, RealFill outpaints the corresponding target images on the right. The region inside the white box is provided to the network as known pixels, and the region outside the white box is generated. RealFill produces high-quality images that are faithful to the references, even when there are dramatic differences between the references and targets such as changes in viewpoint, aperture, lighting, image style, and object motion. 
}
\label{fig:gallery}
\end{figure*}

\begin{figure*}
\centering
\makebox[\textwidth][c]{\includegraphics[width=1\textwidth]{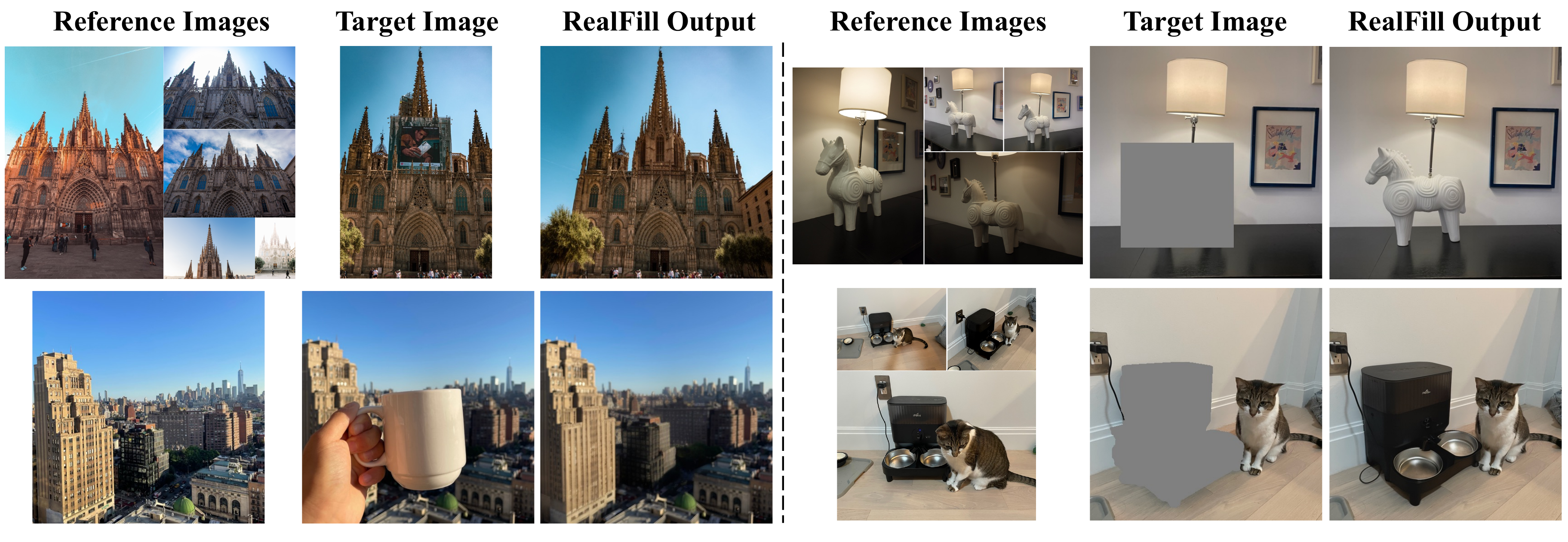}}
\vspace{-0.75cm}
\caption{
\textbf{Reference-based inpainting with RealFill}. Given the references on the left, RealFill can not only remove undesired objects in the target image and reveal the 
occluded contents faithfully (left column), but also insert objects into the scene despite significant viewpoint changes between reference and target images (right column). In the bottom left example, the reference and target images have different defocus blurs. RealFill not only recovers the buildings behind the mug, but also keeps the same amount of blur as in the target image.}
\label{fig:gallery_inpaint}
\end{figure*}

\textbf{Qualitative Results}.
In Figs.~\ref{fig:gallery} and~\ref{fig:gallery_inpaint}, we show that RealFill is able to convincingly outpaint and inpaint image content that is faithful to the reference images. Notably, it is able to handle dramatic differences in camera pose, lighting, defocus blur, image style, and even subject pose. This is because RealFill has both a good image prior (from the pretrained diffusion model) and knowledge of the scene (from finetuning on the input images).

\noindent\textbf{Evaluation Dataset}.
Existing benchmarks for reference-based image completion~\cite{zhou2021transfill} primarily focus on inpainting small regions, and assume at most very minor changes between the 
reference and target images. To better evaluate our target use cases, we create our own dataset, \emph{RealBench}. It consists of 33 scenes (23 outpainting and 10 inpainting), where each scene has a set of reference images $\mathcal{X}_{ref}$, a target image $I_{tgt}$ to fill, a binary mask $M_{tgt}$ indicating the missing region, and the ground-truth result $I_{gt}$. The number of reference images in each scene varies from 1 to 5. The dataset contains diverse, challenging scenarios with significant variations between the reference and target images, such as changes in viewpoint, defocus blur, lighting, style, and subject pose.

\noindent\textbf{Evaluation Metrics}.
We use multiple metrics to evaluate the quality and fidelity of our model outputs. We compare the generated images with the ground-truth target image at multiple levels of image similarity, including PSNR, SSIM, LPIPS~\cite{zhang2018perceptual} for low-level, DreamSim~\cite{fu2023dreamsim} for mid-level, and DINO~\cite{caron2021emerging}, CLIP~\cite{radford2021learning} for high-level. 
For low-level metrics, we only do calculation inside the filled-in region, i.e., where $M_{tgt}$ is 1. For high-level image similarity, we use the cosine distance between the full image embeddings from CLIP and DINO. 
DreamSim~\cite{fu2023dreamsim} is a mid-level similarity between two full images, emphasizing differences in image layouts, object poses, and semantic contents.

\noindent\textbf{Baseline Methods}. We compare to two groups of baselines: models that take reference image as conditioning input, i.e., TransFill~\cite{zhou2021transfill} and Paint-by-Example~\cite{yang2023paint}; and prompt-based image filling approaches including Stable Diffusion Inpainting~\cite{sd2inpaint} and Photoshop Generative Fill~\cite{adobephotoshop}. 
Since TransFill and Paint-by-Example can only use one reference image during inference, we randomly sample one from $\mathcal{X}_{ref}$ as reference
for each run of them. 
Choosing an appropriate prompt for prompt-based filling methods is a necessary component of getting a high quality result. So, for a fair comparison, instead of using a generic prompt like ``a beautiful photo", for each scene, we manually design a long prompt that describe the true scene in detail with the help of ChatGPT~\cite{openai_chatgpt}.

\begin{table}
\caption{
\textbf{Quantitative comparison of RealFill and baselines}.
On RealBench, 
our evaluation set of 33 diverse challenging scenes, RealFill outperforms both prompt-based and reference-based baselines by a large margin on all types of metrics.
}
\vspace{-0.2cm}
\centering
\resizebox{0.475\textwidth}{!}{
\setlength{\tabcolsep}{2.5pt} 
\begin{tabular}{clccc|c|cc}
\toprule
&\multirow{2}{*}{\textbf{Method}}& \multicolumn{3}{c}{low-level}& \multicolumn{1}{c}{mid-level}& \multicolumn{2}{c}{high-level}\\  
\cmidrule{3-8}
&&PSNR$\uparrow$&SSIM$\uparrow$&LPIPS$\downarrow$&DreamSim$\downarrow$&DINO$\uparrow$&CLIP$\uparrow$\\
\midrule
\multirow{2}{*}{\shortstack[c]{prompt\\based}} & SD Inpaint
& 10.63 & 0.282 & 0.605 & 0.213 & 0.831 & 0.874 \\
& Generative Fill
& 10.92 & 0.311 & 0.598 & 0.212 & 0.851 & 0.898 \\
\midrule
\multirow{3}{*}{\shortstack[c]{reference\\based}} & Paint-by-Example
& 10.13 & 0.244 & 0.642 & 0.237 & 0.797 & 0.859 \\
& TransFill
& 13.28 & 0.404 & 0.542 & 0.192 & 0.860 & 0.866 \\
& \textbf{RealFill (ours)} & \textbf{14.78} & \textbf{0.424} & \textbf{0.431} & \textbf{0.077} & \textbf{0.948} & \textbf{0.962} \\
\bottomrule
\end{tabular}
}
\label{tab:quant}
\end{table}

\begin{figure*}
\centering
\makebox[\textwidth][c]{\includegraphics[width=1\textwidth]{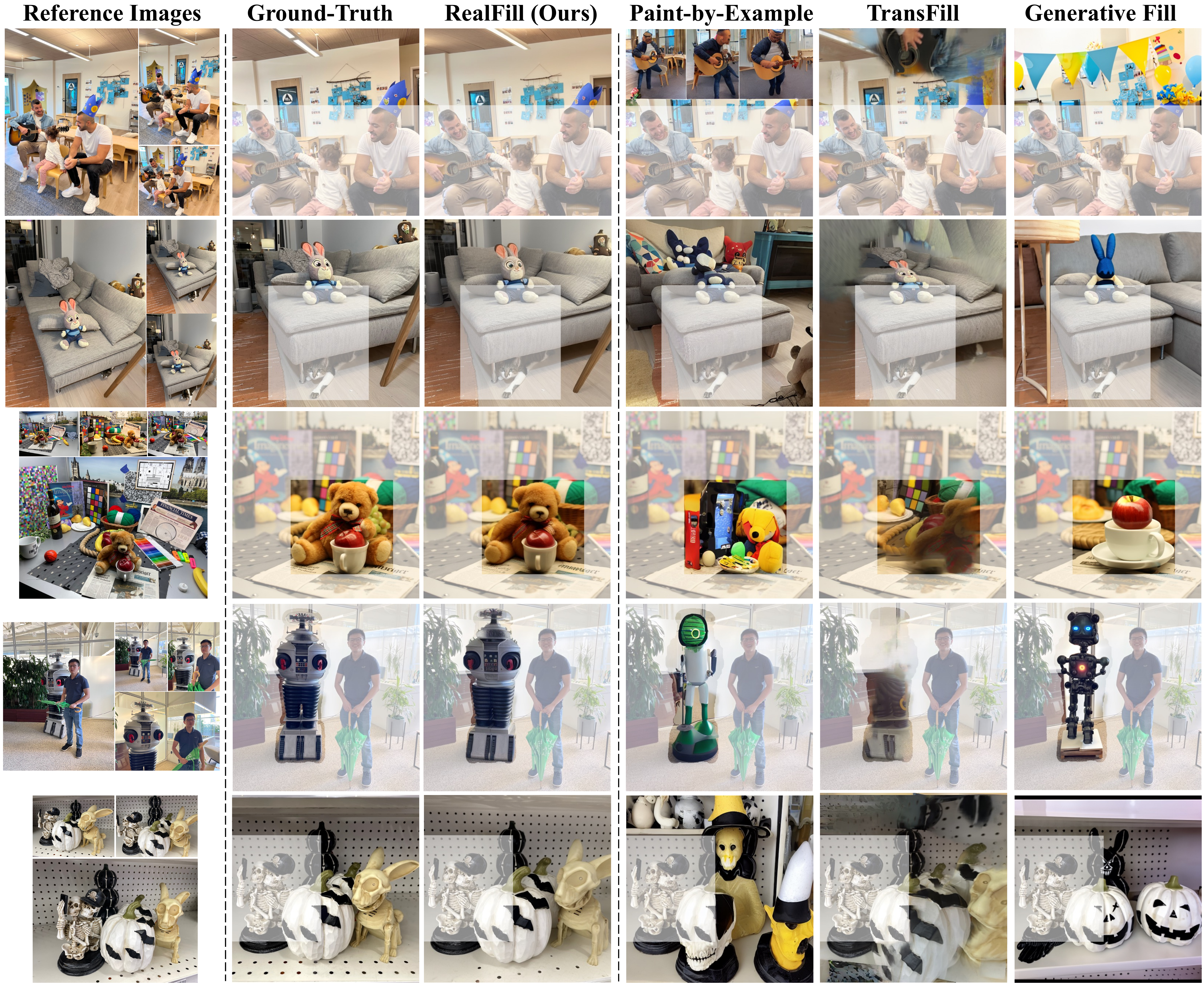}}
\vspace{-0.55cm}
\caption{
\textbf{Qualitative comparison of RealFill and baselines}. Transparent white masks are overlayed on the unaltered known regions of the target images. Paint-by-Example loses fidelity with the reference images because it relies on CLIP embeddings, which only capture high-level semantic information. TransFill outputs low quality images due to the lack of a good image prior and the limitations of its geometry-based pipeline. While Generative Fill produces plausible results, they are inconsistent with the reference images because prompts have limited expressiveness. In contrast, RealFill generates high-quality results that have high fidelity with respect to the reference images. 
}
\label{fig:comparison}
\end{figure*}

\noindent\textbf{Quantitative Comparison}.
We quantitatively evaluate all methods on RealBench. For each approach, we report average metrics across all 33 scenes. Specifically, for generative methods, each scene's metric is itself computed from an average of 64 randomly generated samples (18 for Generative Fill due to Photoshop UI-only limits); for TransFill, the metrics are averaged over different choices of single reference image. 
As shown in Tab.~\ref{tab:quant}, RealFill outperforms all baselines by a large margin across all metrics.

\noindent\textbf{Qualitative Comparison}. 
In Fig.~\ref{fig:comparison}, we present a visual comparison between RealFill and the baselines. We also show the ground-truth and input images for each example. In order to better highlight the regions which are being generated, we overlay a semi-transparent white mask on the ground-truth and output images, covering the known regions of the target image.
RealFill not only generates high-quality images, but also more faithfully reproduces the scene than the baseline methods.
Paint-by-Example relies on the CLIP embedding of the reference images as the condition. This poses a challenge when dealing with complex scenes or attempting to restore object details, since CLIP embeddings only capture high-level semantic information.
Although the geometric-based TransFill has decent numbers in terms of low-level metrics like PSNR, the outputs have much lower quality due to the lack of a good image prior, especially when the scene structure has complex depth variants beyond a planar surface, which is hard for homography transformations to approximate. 
The generated results from Generative Fill are plausible on their own. However, because natural language is limited in conveying complex visual information, they often exhibit substantial deviations from the original scenes depicted in the references.

\begin{table}
\caption{
\textbf{Effect of correspondence-based seed selection}.
It helps RealFill output higher quality results, i.e., filtering out samples with fewer matches results in better quantitative scores.
}
\vspace{-0.2cm}
\centering
\resizebox{.475\textwidth}{!}{
\begin{tabular}{c|cccccc}
\toprule
\multirow{2}{*}{\shortstack[c]{\textbf{Filtering}\\\textbf{Rate}}}&\multirow{2}{*}{PSNR$\uparrow$}&\multirow{2}{*}{SSIM$\uparrow$}&\multirow{2}{*}{LPIPS$\downarrow$}&\multirow{2}{*}{DreamSim$\downarrow$}&\multirow{2}{*}{DINO$\uparrow$}&\multirow{2}{*}{CLIP$\uparrow$}\\
&&&&&&\\
\midrule
0\%&14.78&0.424&0.431&0.077&0.948&0.962\\
25\%&15.01&0.427&0.421&0.066&0.955&0.967\\
50\%&15.05&0.427&0.418&0.063&0.958&0.969\\
\midrule
\textbf{75\%}&\textbf{15.10}&\textbf{0.427}&\textbf{0.417}&\textbf{0.060}&\textbf{0.961}&\textbf{0.970}\\
\bottomrule
\end{tabular}}
\vspace{-0.5cm}
\label{tab:corres_select}
\end{table}

\begin{figure*}
\centering
\makebox[\textwidth][c]{\includegraphics[width=1\textwidth]{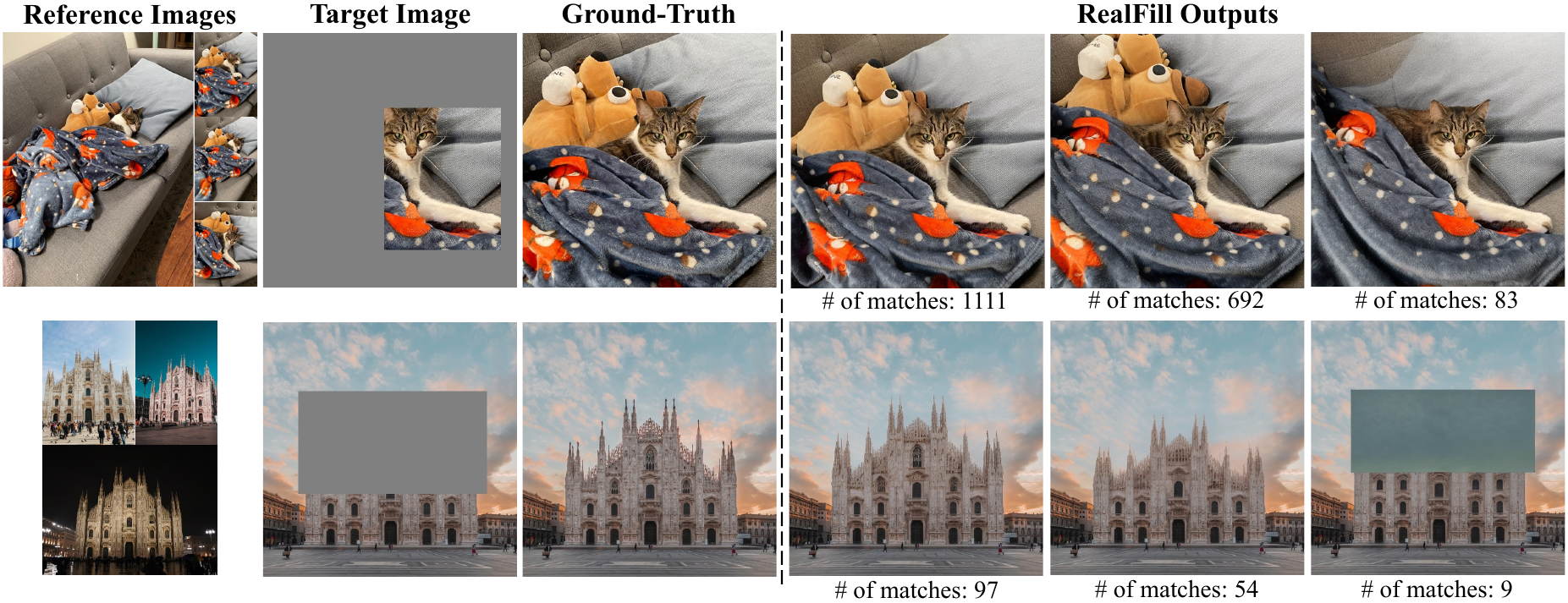}}
\vspace{-0.6cm}
\caption{
\textbf{Correspondence-based seed selection}.
Given the reference images on the left, we show multiple RealFill outputs on the right along with the number of matched key points. We can see that fewer matches correlate with lower-quality outputs that are more divergent from the ground-truth.
}
\label{fig:corres}
\end{figure*}

\begin{figure}
\centering
\includegraphics[width=0.47\textwidth]{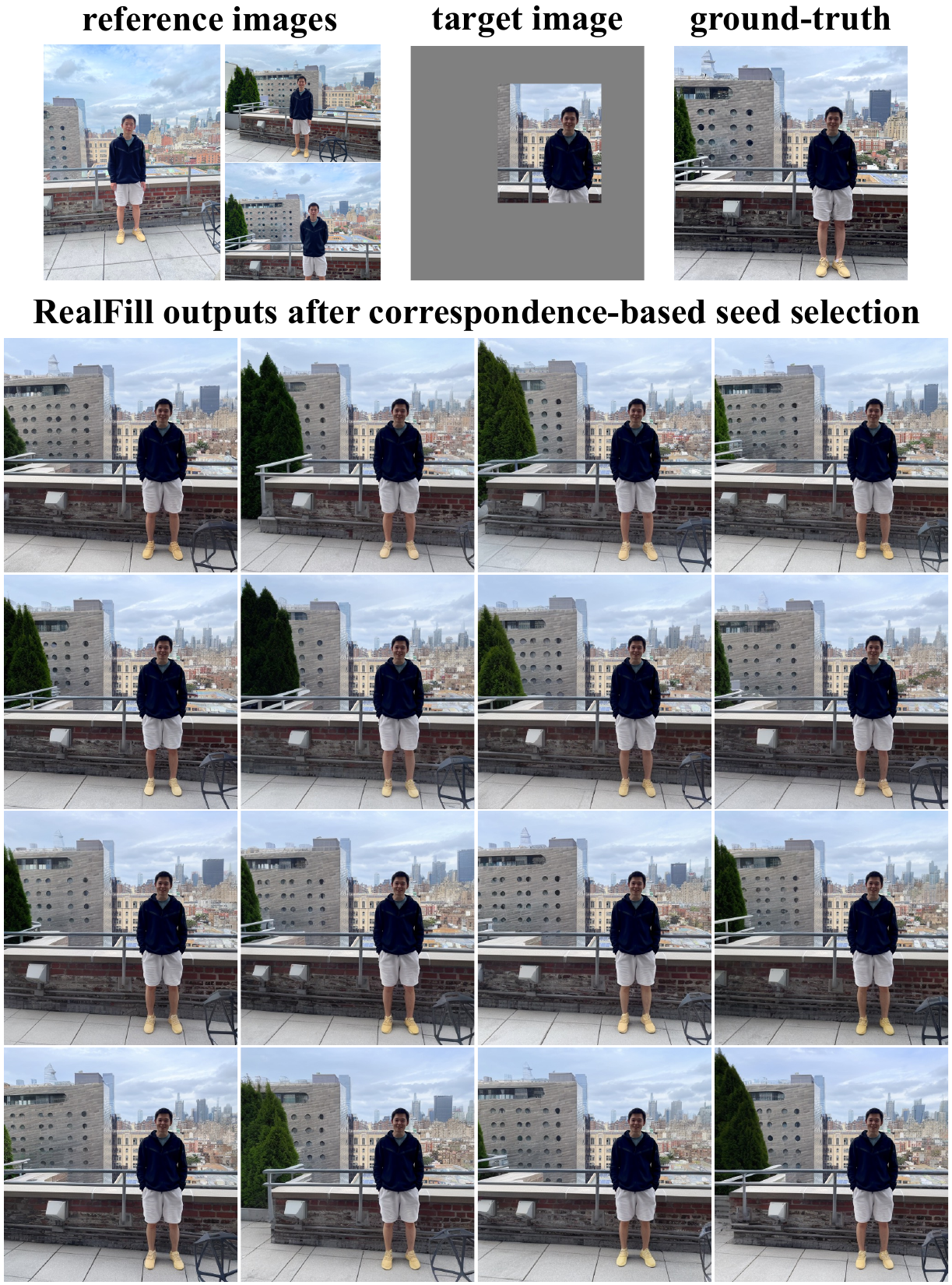}
\vspace{-0.28cm}
\caption{\textbf{Top 16 RealFill outputs after correspondence-based selection}. Model first generates 64 examples in a batch with different sampled noises, then the top 16 images are automatically selected based on predicted correspondences. The outputs are mostly high-quality, so it's easy for users to pick the final image based on their own preference.
}
\label{fig:top16}
\end{figure}

\begin{table}
\caption{
\textbf{User study results}.
Here we show the preference percentages for the most realistic and most faithful image completions across 58 scenes by 44 participants. RealFill significantly surpasses baselines, especially on the faithfulness criterion.
}
\vspace{-0.2cm}
\centering
\resizebox{0.475\textwidth}{!}{
\begin{tabular}{lc|c}
\toprule
\textbf{Method} & \textbf{Most Realistic} $\uparrow$ & \textbf{Most Faithful} $\uparrow$ \\
\midrule
TransFill~\cite{zhou2021transfill}& 2.0\% & 3.7\% \\
Paint-by-Example~\cite{yang2023paint}& 11.0\% & 1.9\% \\
Generative Fill~\cite{adobephotoshop}& 23.4\% & 7.2\% \\
\midrule
\textbf{RealFill (ours)} & \textbf{63.7\%} & \textbf{87.2\%} \\
\bottomrule
\end{tabular}
}
\label{tab:study}
\end{table}

\noindent\textbf{Correspondence-Based Seed Selection}. 
We evaluate the effect of our proposed correspondence-based seed selection described in Sec.~\ref{sec:realfill}. To measure the correlation between our seed selection mechanism and high-quality results, for each scene, we rank RealFill's outputs $\{I_{out}\}$ according to the number of matched keypoints, and then filter out a certain percent of the lowest-ranked samples. We then average the evaluation metrics only across the remaining samples. Higher filtering rates like 75\% are quantitatively better than no filtering (Tab.~\ref{tab:corres_select}).
In Fig~\ref{fig:corres}, we show multiple RealFill outputs with the corresponding number of matched keypoints. These demonstrate a clear trend, where fewer matches usually indicate lower-quality results.

Therefore, we followed such strategy to select RealFill outputs in \cref{fig:teaser,fig:method,fig:gallery,fig:gallery_inpaint,fig:comparison}: for each scene, model first generates 64 examples in a batch with different sampled noises, then we only keep the top 16 images based on predicted correspondences, and manually pick one from them. For fair comparison, we also manually pick the best outputs for each baseline as in \cref{fig:teaser,fig:comparison}.
As shown in \cref{fig:top16}, after correspondence-based selection, the outputs are mostly high-quality with small variations. Users can make choices based on their own preference, therefore the involved human labor is very light.

\noindent\textbf{User study}. 
In our study, 44 users evaluated the realism and faithfulness of image completions from four different methods across 58 scenes. Participants first review an incomplete target image alongside the outputs of these methods, choosing the one they found most realistic. Then, with reference images provided, they select the most faithful completion. The study includes a mix of RealBench and 25 additional challenging scenes. All method outputs are randomly sampled to avoid bias.
Overall, 2552 votes per criterion were collected. The results, as summarized in Tab.~\ref{tab:study}, show RealFill's superiority compared to the baseline methods especially in faithfulness.
\section{Discussion}

\begin{figure}
\centering
\includegraphics[width=0.475\textwidth]{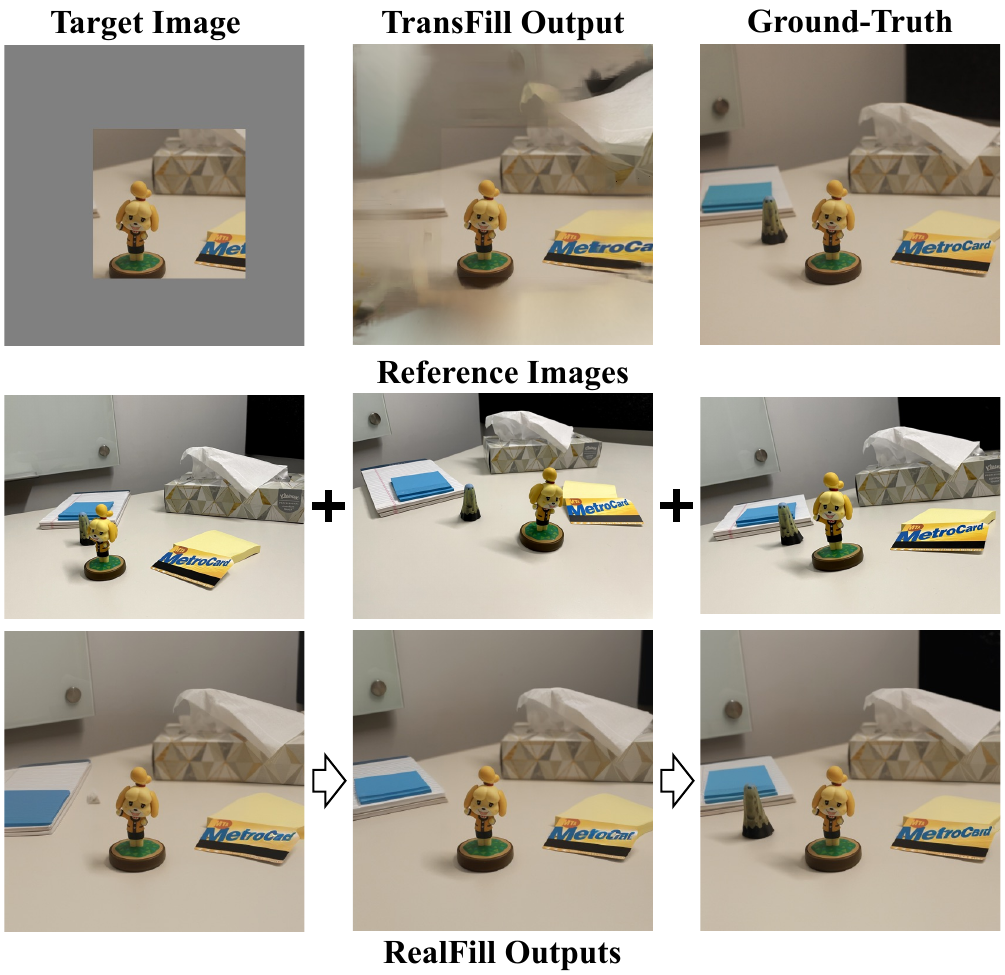}
\vspace{-0.65cm}
\caption{
\textbf{Influence of reference image choice on RealFill}. For the given target image, we show different RealFill outputs by gradually increasing number of reference images from left to right. We can see outputs are getting better when having more references, e.g., the ghost gets recovered when having all three reference images.
Note that all RealFill outputs significantly outperform the TransFill baseline.
}
\label{fig:ablate_ref}
\end{figure}

\noindent\textbf{How does reference image choice affect RealFill?}
Empirically, when there are more reference images, or when the references have smaller variations from the target in terms of viewpoint and lighting, RealFill gives better results, as shown in \cref{fig:ablate_ref}.

\noindent\textbf{Would other baselines work?}

\begin{figure}
\centering
\includegraphics[width=0.475\textwidth]{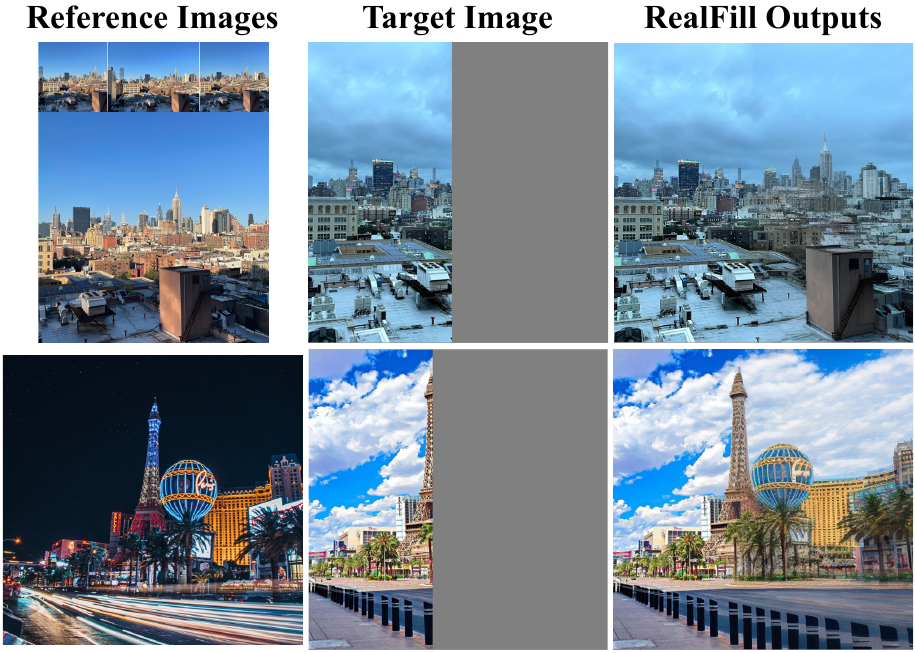}
\vspace{-0.55cm}
\caption{
\textbf{RealFill vs. image stitching}.
Commercial image stitching softwares fail to produce any outputs when there are significant variations between reference and target images, such as disparate lighting conditions. In contrast, RealFill excels by producing accurate and high-quality results. It effectively recovers elements like rooftop tanks and balloons, maintaining fidelity even under varying lighting scenarios between the compared images.
}
\label{fig:hard2stitch}
\end{figure}

\textit{Image Stitching}. 
It is possible to stitch the reference and target images together using correspondences. However, we find that even strong commercial image stitching software doesn't work when there are dramatic lighting changes or object motion. 
Taking the two scenes in Fig.~\ref{fig:hard2stitch} for example, multiple commercial software solutions produce no output, asserting that there are insufficient correspondences. In contrast, RealFill faithfully recovers these scenes.

\begin{figure}
\centering
\includegraphics[width=0.475\textwidth]{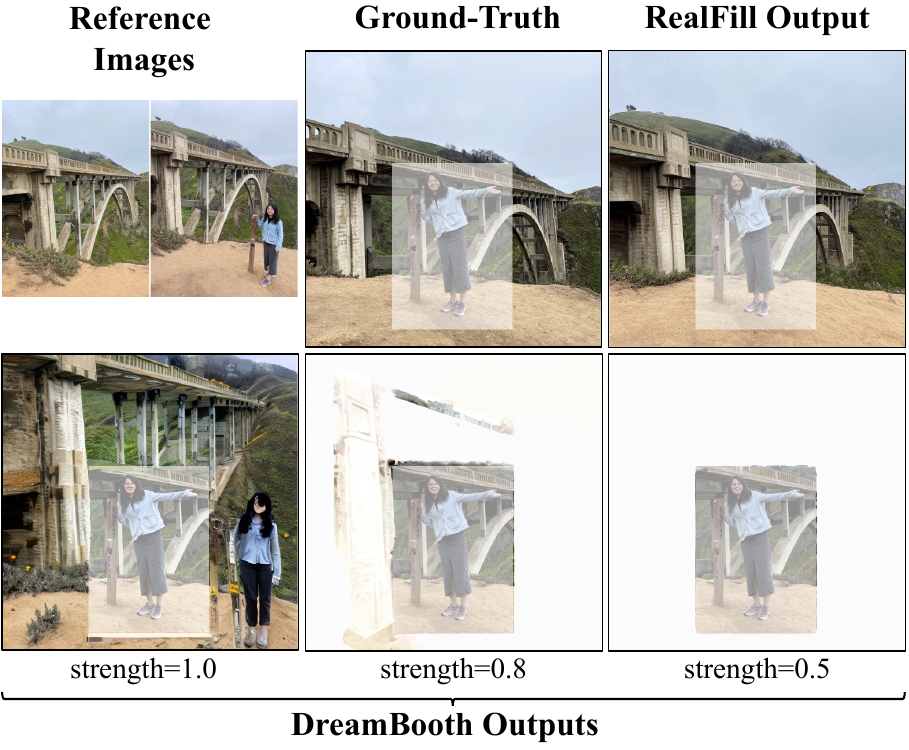}
\vspace{-0.7cm}
\caption{
\textbf{RealFill vs. DreamBooth}.
Finetuning a standard Stable Diffusion model on reference images and using it to fill missing regions, leads to drastically worse results compared to RealFill. We show samples for various levels of the strength hyper-parameter.
}
\label{fig:compare_db}
\end{figure}

\textit{DreamBooth}.
Instead of finetuning an inpainting model, an alternative is to finetune a standard Stable Diffusion model on the reference images, i.e., DreamBooth, then use the finetuned T2I model to inpaint the target image~\cite{lugmayr2022repaint}, as implemented in the popular Diffusers library~\cite{von-platen-etal-2022-diffusers}\footnote{\scriptsize Diffusers' Stable Diffusion inpainting pipeline \href{https://github.com/huggingface/diffusers/blob/19edca82f1ff194c07317369a92b470dbae97f34/src/diffusers/pipelines/stable_diffusion/pipeline_stable_diffusion_inpaint.py}{code}.}. 
However, because this model is never trained with a masked prediction objective, it is much worse than RealFill, as shown in Fig.~\ref{fig:compare_db}.

\begin{figure}
\centering
\includegraphics[width=0.475\textwidth]{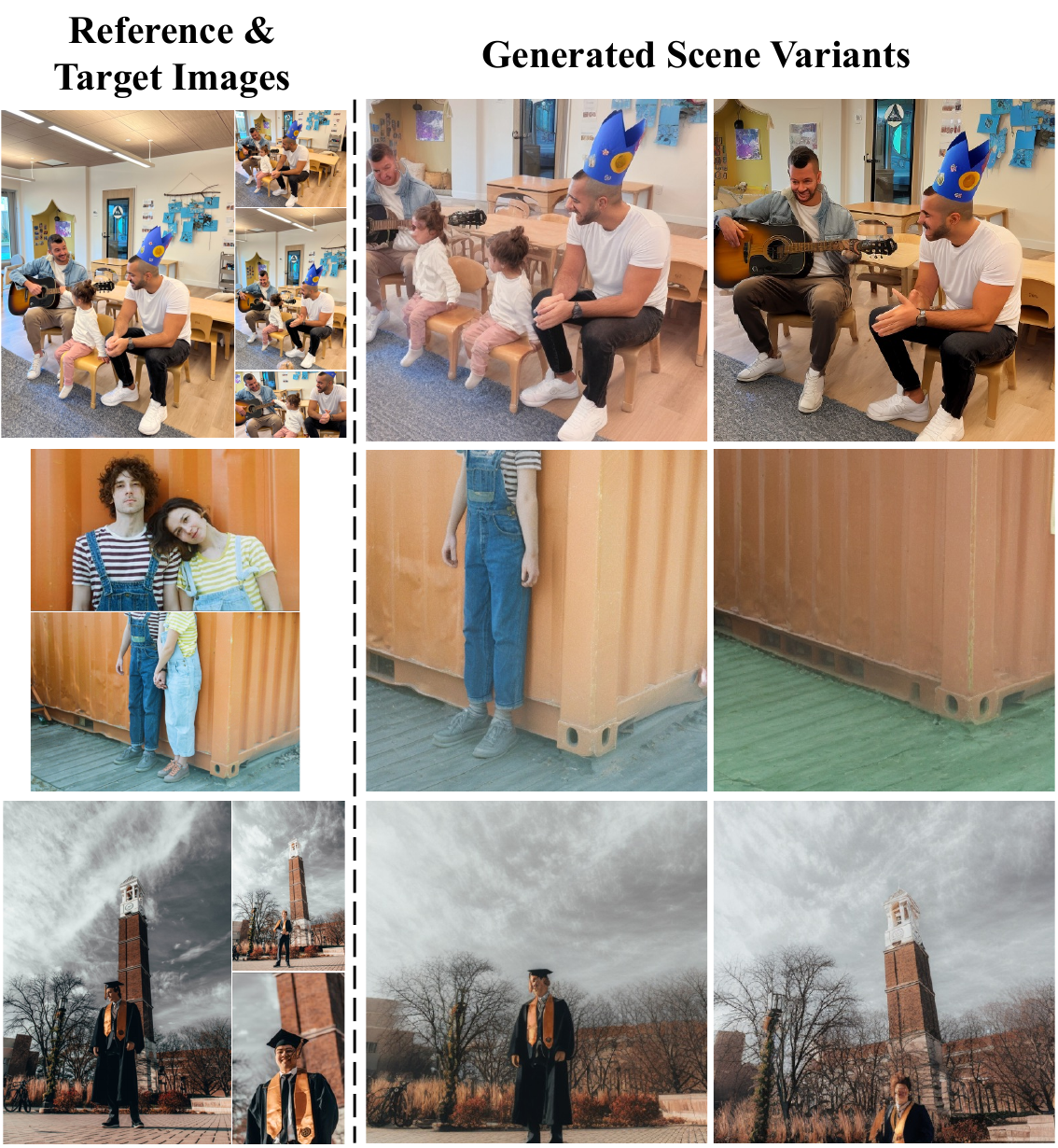}
\vspace{-0.6cm}
\caption{
RealFill is able to generate multiple scene variants when conditioned on a blank image as input, e.g., people are added or removed in the first and second rows. This suggests that the finetuned model can relate elements inside the scene in a compositional manner.
}
\label{fig:null_mask}
\end{figure}

\noindent\textbf{What makes RealFill work?}
To explore why RealFill leads to strong results, especially on complex scenes, we make the following two hypotheses:

\textit{RealFill relates multiple elements in a scene}.
If we make the conditioning image a blank canvas during inference, i.e., all entries of $M_{tgt}$ equal 1, we can see in Fig.~\ref{fig:null_mask} that the finetuned model is able to generate multiple scene variants with different structures, e.g., removing the foreground or background object, or manipulating the object layouts. This suggest that RealFill may understand the scene composition.

\begin{figure}
\centering
\includegraphics[width=0.475\textwidth]{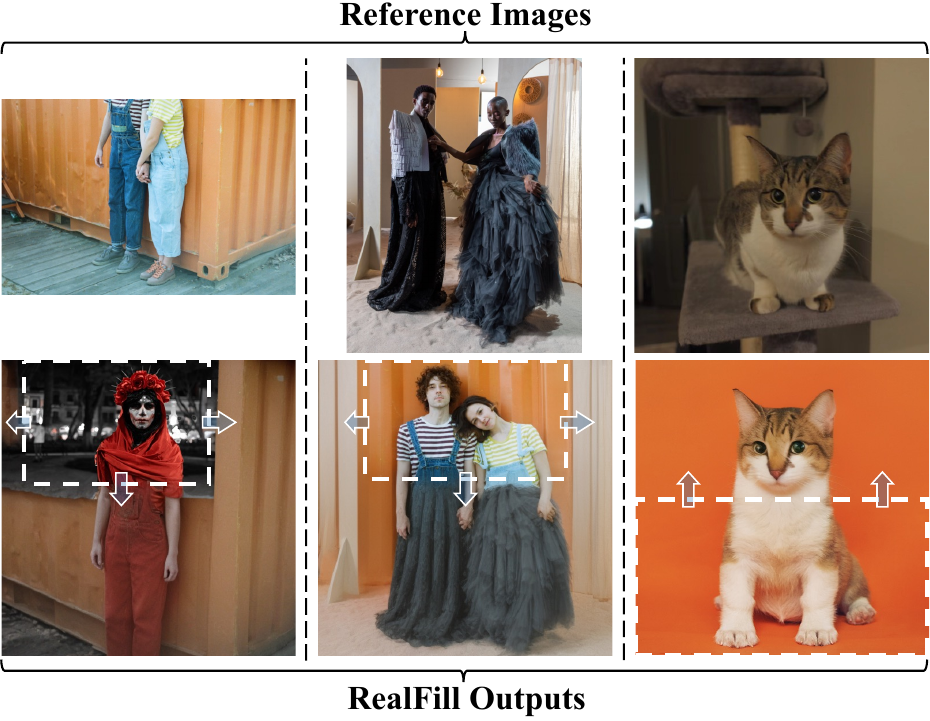}
\vspace{-0.6cm}
\caption{
When the reference and target images do not depict the same scene, the finetuned model is still able to fuse the reference contents into the target image in a semantically-reasonable way, suggesting that it captures both real or invented correspondences between input images.
}
\label{fig:mismatch}
\end{figure}

\textit{RealFill captures correspondences among input images}.
Even if the reference and target images do not depict the same scene, the finetuned model is still able to fuse the corresponding contents of the reference images into the target area seamlessly, as shown in Fig.~\ref{fig:mismatch}. This suggests that RealFill is able to capture and utilize real or invented correspondences between reference and target images to do generation. Previous works~\cite{tang2023dift,luo2023diffusion,zhang2023tale} also found similar emergent correspondence inside pretrained diffusion models.

\begin{figure}
\centering
\includegraphics[width=0.475\textwidth]{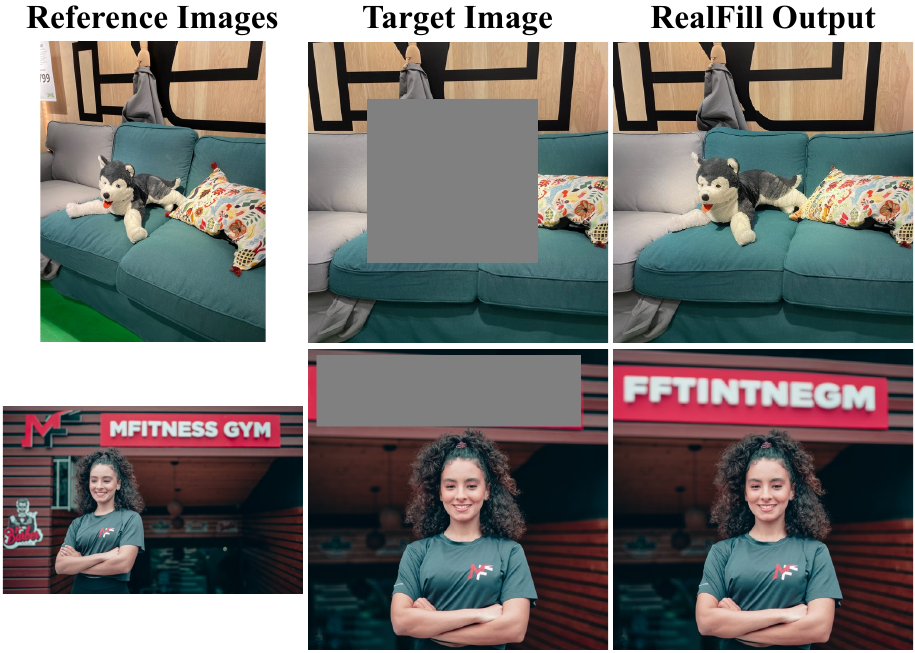}
\vspace{-0.5cm}
\caption{\textbf{Failure cases of RealFill}. 
(Top) RealFill fails to recover the precise 3D scene structure, e.g., the output husky plush has different pose compared to the reference; (Bottom) RealFill fails to handle cases that are also challenging for the base T2I diffusion model, e.g., the words on the store sign are wrongly spelled.
}
\label{fig:failure}
\end{figure}

\noindent\textbf{Limitations}. Because RealFill requires a gradient-based finetuning process on input images, it is relatively slow.
Empirically, we also find that, when there is a large viewpoint change between reference and target images, RealFill fails to recover the 3D scene faithfully, especially when there's only a single reference image, as in Fig.~\ref{fig:failure} top.
In addition, because RealFill relies on the base pretrained model's image prior, it also fails to handle cases that are challenging for the base model, e.g., Stable Diffusion is not good at generating fine details, such as text, human faces, or body parts, as in Fig.~\ref{fig:failure} bottom.

Lastly, similar to the user study in Tab.~\ref{tab:study}, we conducted another study on RealBench where participants compare randomly sampled RealFill output vs. ground-truth image for each scene. Among the collected 396 pairwise comparisons, RealFill only gets 23.7\% of the votes for realism and 22.0\% for faithfulness, vs. ground-truth's 76.3\% and 78.0\% respectively. This is reasonable because it's easy for human to spot artifacts especially for side-by-side comparison, but it also shows that more future improvements are needed to make RealFill achieve perfect authentic image completion.

\noindent\textbf{Societal Impact}.
This research aims to create a tool that can help users express their creativity and improve the quality of their personal photographs through image generation. However, advanced image generation methods can have complex impacts on society. Our proposed method inherits some of the concerns that are associated with this class of technology, such as the potential to alter sensitive personal characteristics. The open-source pretrained model that we use in our work, Stable Diffusion, exhibits some of these concerns. However, we have not found any evidence that our method is more likely to produce biased or harmful content than previous work. Despite these findings, it is important to continue investigating the potential risks of image generation technology. Future research should focus on developing methods to mitigate bias and harmful content, and to ensure that image generation tools are used in a responsible manner.
\section{Conclusion}

In this work, we introduce the problem of \textit{\Adjective{} Image Completion}, where given a few reference images, we intend to complete some missing regions of a target image with the content that ``\emph{should} have been there'' rather than ``what \emph{could} have been there''. To tackle this problem, we proposed a simple yet effective approach called RealFill, which first finetunes a T2I inpainting diffusion model on the reference and target images, and then uses the adapted model to fill the missing regions. We show that RealFill produces high-quality image completions that are faithful to the content in the reference images, even when there are large differences between reference and target images such as viewpoint, aperture, lighting, image style, and object pose.

\begin{acks}
Luming and Bharath are supported by NSF IIS-2144117.
We would like to thank Rundi Wu, Qianqian Wang, Viraj Shah, Ethan Weber, Zhengqi Li, Kyle Genova, Richard Tucker, Boyang Deng, Maya Goldenberg, Noah Snavely, Ben Poole, Ben Mildenhall, Alex Rav-Acha, Pratul Srinivasan, Dor Verbin, Jon Barron and all the anonymous reviewers for their valuable discussion and feedbacks, and thank Zeya Peng, Rundi Wu, Shan Nan for their contribution to the evaluation dataset. A special thanks to Jason Baldridge, Kihyuk Sohn, Kathy Meier-Hellstern, and Nicole Brichtova for their feedback and support for the project.
\end{acks}
\appendix



\section{Implementation Details}

\subsection{RealFill}
For each scene, we finetune the Stable Diffusion inpainting model~\cite{sd2inpaint} for 2,000 iterations with a batch size of 16 on a single NVIDIA A100 GPU with LoRA rank 8. With a probability of 0.1, we randomly dropout prompt $p$, mask $m$ and LoRA layers independently during training. The learning rate is set to 2e-4 for the U-Net and 4e-5 for the text encoder. The whole finetuning process takes around one hour but usually 20 minutes would give pretty good results already in many cases. Note that these hyper-parameters could be further tuned for each scene to get better performance, e.g., some scenes converge more quickly may overfit if trained for too long. However, for the sake of fair comparison, we use a constant set of hyper-parameters for all results shown in the paper. During inference, we use DDPM~\cite{ho2020denoising}
sampler with step 200 and guidance weight 1.0, i.e., without classifier-free guidance.

\subsection{User Study}

\begin{figure}[h]
\centering
\includegraphics[width=0.475\textwidth]{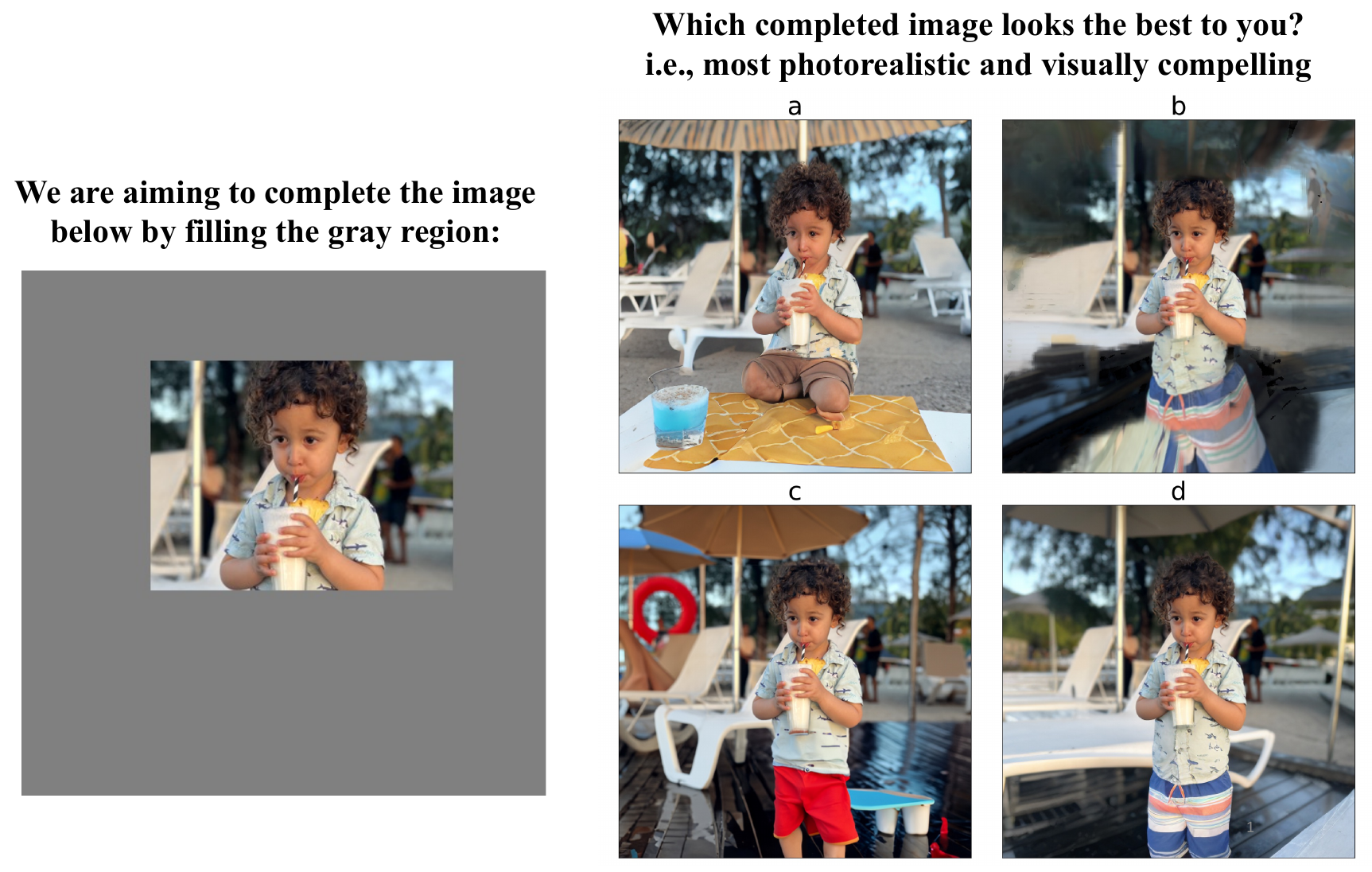}
\vspace{-0.55cm}
\caption{An example of the realism question in the user study. Users first review an incomplete target image alongside the shuffled outputs of four methods, then choose the one they find most realistic.}
\label{fig:user_a}
\end{figure}

\begin{figure}
\centering
\includegraphics[width=0.475\textwidth]{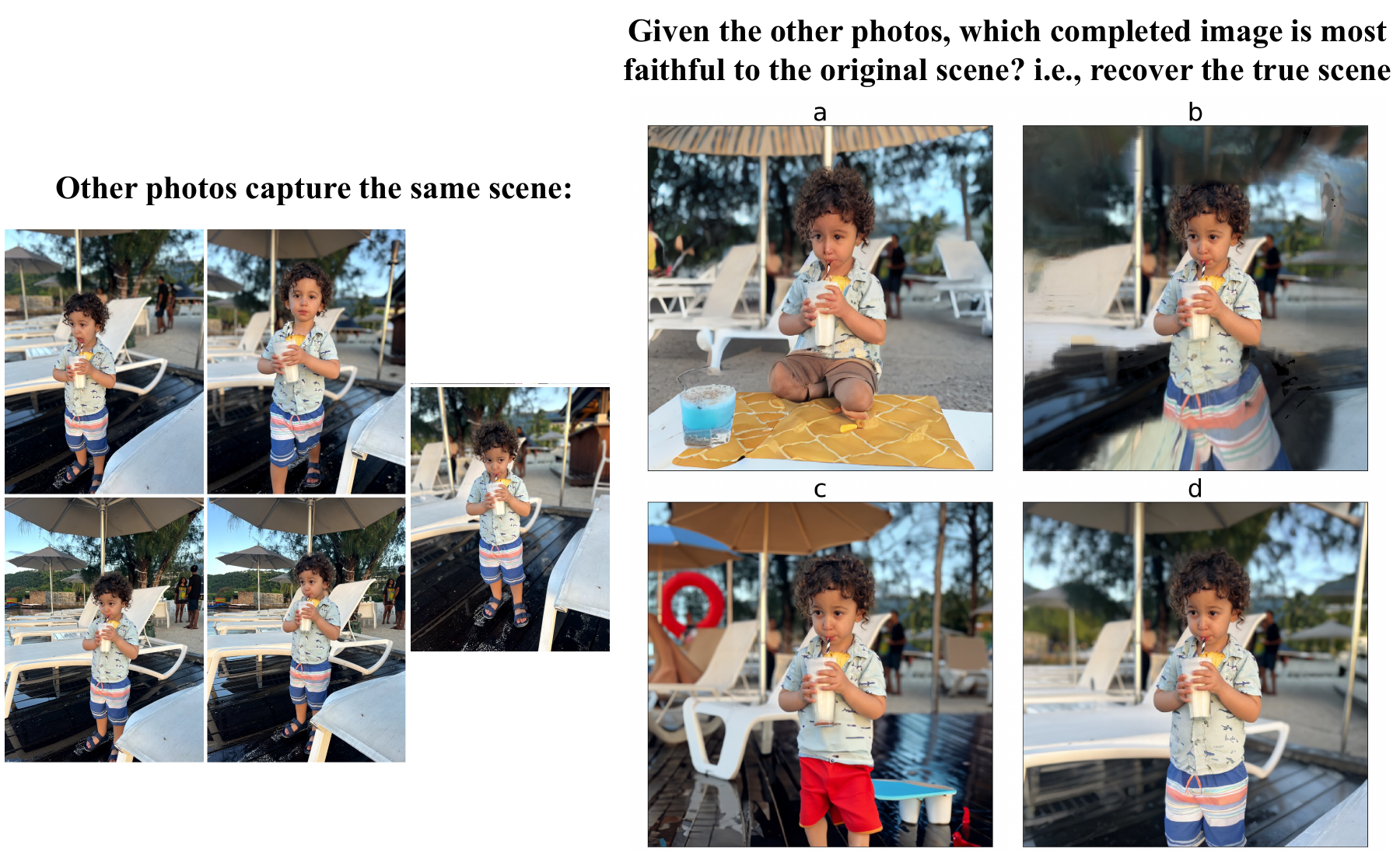}
\vspace{-0.5cm}
\caption{An example of the faithfulness question in the user study. After the user answered the realism question, the reference images are provided. Then, they are asked to select the most faithful completion.}
\label{fig:user_b}
\end{figure}

In the user study, participants are asked to evaluated the realism and faithfulness of image completions across 58 scenes from four different methods including RealFill, Paint-by-Example~\cite{yang2023paint}, TransFill~\cite{zhou2021transfill}, and Photoshop Generative Fill~\cite{adobephotoshop}. All method outputs are randomly sampled without human intervention. For each scene, the placement organization of different method outputs are also randomly shuffled to avoid user bias. 

Specifically, participants first review an incomplete target image alongside the outputs of these methods, choosing the one they found most realistic, as shown in Fig.~\ref{fig:user_a}. Then, with reference images provided, they select the most faithful completion, as shown in Fig.~\ref{fig:user_b}. 

\bibliographystyle{ACM-Reference-Format}
\bibliography{egbib}



\end{document}